\documentclass{article}
\pdfoutput=1
\usepackage[margin=1.42in]{geometry}
\PassOptionsToPackage{numbers}{natbib}
\usepackage[utf8]{inputenc} 
\usepackage[T1]{fontenc}    
\usepackage{hyperref}       
\usepackage{url}            
\usepackage{booktabs}       
\usepackage{amsfonts}       
\usepackage{nicefrac}       
\usepackage{microtype}      
\usepackage{algorithm2e}
\usepackage{subcaption}
\usepackage{fancyvrb}
\usepackage{color}
\usepackage{xspace}
\usepackage{url}
\usepackage{listings}
\usepackage[toc,page]{appendix}

\usepackage{listings}

\definecolor{framegray}{rgb}{0.4,0.4,0.4}

\lstdefinestyle{frameStyle}{
  language=c++,
  numbers=left,
  stepnumber=1,
  numbersep=10pt,
  tabsize=4,
  showspaces=false,
  showstringspaces=false
  numberstyle=\tiny,      
  basicstyle=\scriptsize \sffamily,    
  xleftmargin=2.5em,
  frame=single,
  keywordstyle=\color{blue}\sffamily,
  stringstyle=\color{red}\sffamily,
  commentstyle=\color{framegray}\sffamily,
  morecomment=[l][\color{magenta}]{\#}
}

\newcommand{\starcraft}{StarCraft: Brood War\xspace}

\title{TorchCraft: a Library for Machine Learning Research\\
on Real-Time Strategy Games}

\author{
  Gabriel Synnaeve, Nantas Nardelli, Alex Auvolat, Soumith Chintala,\\ Timoth\'{e}e Lacroix, Zeming Lin, Florian Richoux, Nicolas Usunier\\
  \texttt{gab@fb.com, nantas@robots.ox.ac.uk}
}

\begin{document}

\maketitle


\begin{abstract}

  We present TorchCraft, a library that enables deep learning
    research on Real-Time Strategy (RTS) games such as \starcraft, by making it
    easier to control these games from a machine learning framework, here
    Torch~\cite{torch7}. This white paper argues for using RTS games as a
    benchmark for AI research, and describes the design and components of
    TorchCraft.
  
\end{abstract}



\section{Introduction}

Deep Learning techniques \cite{lecun2015deep} have recently enabled researchers
to successfully tackle low-level perception problems in a supervised learning
fashion. In the field of Reinforcement Learning this has transferred into the
ability to develop agents able to learn to act in high-dimensional input
spaces. In particular, deep neural networks have been used to help
reinforcement learning scale to environments with visual inputs, allowing them
to learn policies in testbeds that previously were completely intractable.  For
instance, algorithms such as Deep Q-Network (DQN) \cite{mnih2015human} have
been shown to reach human-level performances on most of the classic ATARI 2600
games by learning a controller directly from raw pixels, and without any
additional supervision beside the score. Most of the work spawned in this new
area has however tackled environments where the state is fully observable, the
reward function has no or low delay, and the action set is relatively small. To
solve the great majority of real life problems agents must instead be able to
handle partial observability, structured and complex dynamics, and noisy and
high-dimensional control interfaces.


To provide the community with useful research environments, work was done
towards building platforms based on videogames such as Torcs
\cite{wymann2000torcs}, Mario AI \cite{togelius20102009}, Unreal's BotPrize
\cite{hingston2009turing}, the Atari Learning Environment
\cite{bellemare2012arcade}, VizDoom \cite{kempka2016vizdoom}, and Minecraft
\cite{johnson2016malmo}, all of which have allowed researchers to train deep
learning models with imitation learning, reinforcement learning and various
decision making algorithms on increasingly difficult problems. Recently there
have also been efforts to unite those and many other such environments in one
platform to provide a standard interface for interacting with them
\cite{brockman2016openai}. We propose a bridge between StarCraft: Brood War, an
RTS game with an active AI research community and annual AI competitions
\cite{ontanon2013survey,scaicomp,bwapi}, and Lua, with examples in Torch
\cite{torch7} (a machine learning library).


\section{Real-Time Strategy for Games AI}

Real-time strategy (RTS) games have historically been a domain of interest of
the planning and decision making research communities
\cite{buro2004rts,aha2005learning,scaicomp,ontanon2013survey,robertson2014review}.
This type of games aims to simulate the control of multiple units in a military
setting at different scales and level of complexity, usually in a fixed-size 2D
map, in duel or in small teams.  The goal of the player is to collect resources
which can be used to expand their control on the map, create buildings and
units to fight off enemy deployments, and ultimately destroy the opponents.
These games exhibit durative moves (with complex game dynamics) with
simultaneous actions (all players can give commands to any of their units at
any time), and very often partial observability (a ``fog of war'': opponent 
units not in the vicinity of a player's units are not shown).

\textbf{RTS gameplay: }
Components RTS game play are economy and battles (``macro'' and ``micro''
respectively): players need to gather resources to build military units and
defeat their opponents. To that end, they often have worker units (or
extraction structures) that can gather resources needed to build workers,
buildings, military units and research upgrades. Workers are often also
builders (as in StarCraft), and are weak in fights compared to military units.
Resources may be of varying degrees of abundance and importance. For instance,
in StarCraft minerals are used for everything, whereas gas is only required for
advanced buildings or military units, and technology upgrades. Buildings and
research define technology trees (directed acyclic graphs) and each state of a
``tech tree'' allow for the production of different unit types and the training
of new unit abilities. Each unit and building has a range of sight that
provides the player with a view of the map. Parts of the map not in the sight
range of the player's units are under fog of war and the player cannot observe
what happens there. A considerable part of the strategy and the tactics lies in
which armies to deploy and where.

Military units in RTS games have multiple properties which differ between unit
types, such as: attack range (including melee), damage types, armor, speed,
area of effects, invisibility, flight, and special abilities. Units can have
attacks and defenses that counter each others in a rock-paper-scissors fashion,
making planning armies a extremely challenging and strategically rich process.
An ``opening'' denotes the same thing as in Chess: an early game plan for which
the player has to make choices. That is the case in Chess because one can move
only one piece at a time (each turn), and in RTS games because, during the
development phase, one is economically limited and has to choose 
which tech paths to pursue. Available resources constrain the technology
advancement and the number of units one can produce. As producing buildings
and units also take time, the arbitrage between investing in the economy, in
technological advancement, and in units production is the crux of the strategy
during the whole game.


\textbf{Related work: } Classical AI approaches normally involving planning and
search \cite{aha2005learning, ontanon2007case, weber2014reactive,
  churchill2016heuristic} are extremely challenged by the combinatorial action
space and the complex dynamics of RTS games, making simulation (and thus Monte
Carlo tree search) difficult \cite{churchill2012fast, uriarte2014game}. Other
characteristics such as partial observability, the non-obvious quantification of
the value of the state, and the problem of featurizing a dynamic and structured
state contribute to making them an interesting problem, which altogether
ultimately also make them an excellent benchmark for AI. As the scope of this
paper is not to give a review of RTS AI research, we refer the reader to these
surveys about existing research on RTS and StarCraft AI \cite{ontanon2013survey,
  robertson2014review}.


It is currently tedious to do machine learning research in this
domain. Most previous reinforcement learning research involve simple models or
limited experimental settings \cite{wender2012applying,usunier2016episodic}.
Other models are trained on offline datasets of highly skilled players
\cite{weber2009data,synnaeve2012bayesian,synnaeve2012dataset,novadata}.
Contrary to most Atari games \cite{bellemare2012arcade}, RTS games have much
higher action spaces and much more structured states. Thus, we advocate here
to have not only the pixels as input and keyboard/mouse for commands, as in
\cite{bellemare2012arcade,brockman2016openai,kempka2016vizdoom}, but also a
structured representation of the game state, as in \cite{johnson2016malmo}.
This makes it easier to try a broad variety of models, and may be useful in
shaping loss functions for pixel-based models.

Finally, StarCraft: Brood War is a highly popular game (more than 9.5 million
copies sold) with professional players, which provides interesting datasets,
human feedback, and a good benchmark of what is possible to achieve within the
game. There also exists an active academic community that organizes AI
competitions.


\section{Design}


The simplistic design of TorchCraft is applicable to any video game and any
machine learning library or framework. Our current implementation connects
Torch to a low level interface \cite{bwapi} to StarCraft: Brood War.
TorchCraft's approach is to dynamically inject a piece of code in the game
engine that will be a server. This server sends the state of the game to a
client (our machine learning code), and receives commands to send to the game.
This is illustrated in Figure~\ref{fig:loops}.
The two modules are entirely synchronous, but the we provide two modalities of
execution based on how we interact with the game:

\begin{description}
\item[Game-controlled] - we inject a DLL that provides the game interface to the
  bots, and one that includes all the instructions to communicate with the
  machine learning client, interpreted by the game as a player (or bot AI). In
  this mode, the server starts at the beginning of the match and shuts down when
  that ends. In-between matches it is therefore necessary to re-establish the
  connection with the client, however this allows for the setting of multiple
  learning instances extremely easily.
\item[Game-attached] - we inject a DLL that provides the game interface to the
  bots, and we interact with it by attaching to the game process and
  communicating via pipes. In this mode there is no need to re-establish the
  connection with the game every time, and the control of the game is completely
  automatized out of the box, however it's currently impossible to create
  multiple learning instances on the same guest OS.
\end{description}

Whatever mode one chooses to use, TorchCraft is seen by the AI programmer as
a library that provides: \texttt{connect()}, \texttt{receive()} (to get the
state), \texttt{send(commands)}, and some helper functions about specifics of
StarCraft's rules and state representation. TorchCraft also provides an
efficient way to store game frames data from past (played or observed) games
so that existing state (``replays'', ``traces'') can be re-examined.

\begin{figure}
  \centering
  \begin{subfigure}{0.45\textwidth}
    \begin{Verbatim}
-- main game engine loop:
while true do
    game.receive_player_actions()
    game.compute_dynamics()
    -- our injected code:
    torchcraft.send_state()
    torchcraft.receive_actions()
end
    \end{Verbatim}
  \end{subfigure}
  \quad \quad
  \begin{subfigure}{0.45\textwidth}
    \begin{Verbatim}
featurize, model = init()
tc = require 'torchcraft'
tc:connect(port)
while not tc.state.game_ended do
    tc:receive()
    features = featurize(tc.state)
    actions = model:forward(features)
    tc:send(tc:tocommand(actions))
end
    \end{Verbatim}
  \end{subfigure}
  \caption{Simplified client/server code that runs in the game engine (server,
    on the left) and the library for the machine learning library or framework
    (client, on the right).}
  \label{fig:loops}
\end{figure}


\section{Conclusion}

We presented several work that established RTS games as a source of interesting
and relevant problems for the AI research community to work on. We believe that
an efficient bridge between low level existing APIs and machine learning
frameworks/libraries would enable and foster research on such games. We
presented TorchCraft: a library that enables state-of-the-art machine learning
research on real game data by interfacing Torch with StarCraft: BroodWar.
TorchCraft has already been used in reinforcement learning experiments on
StarCraft, which led to the results in \cite{usunier2016episodic} (soon to be
open-sourced too and included within TorchCraft).


\section{Acknowledgements}

We would like to thank Yann LeCun, Léon Bottou, Pushmeet Kohli, Subramanian
Ramamoorthy, and Phil Torr for the continuous feedback and help with various
aspects of this work.
Many thanks to David Churchill 
for proofreading early versions of this paper.



\bibliographystyle{acm} 
\renewcommand{\baselinestretch}{0.9}
{\small
\bibliography{main}
}

\pagebreak

\appendix

\section{Frame data}
\label{appendix:frame}

In addition to the visual data, the TorchCraft server extracts certain
information for the game state and sends it over to the connected clients in a
structured ``frame''. The frame is formatted in a table in roughly the following
structure:

\lstset{style=frameStyle}

\begin{lstlisting}
Received update:  {
 // Number of frames in the current game
 // NB: a 'game' can be composed of several battles
 frame_from_bwapi : int
  units_myself :
   {
   // Unit ID
   int:
    {
    // Unit ID
    target : int
    targetpos :
     {
     // Absolute x
     1 : int
     // Absolute y
     2 : int
     }
    // Type of air weapon
    awtype : int 
    // Type of ground weapon
    gwtype : int
    // Number of frames before next air weapon possible attack
    awcd : int
    // Number of hit points
    hp : int
    // Number of energy / mana points, if any
    energy : int
    // Unit type
    type : int 
    position :
     {
     // Absolute x
     1 : int
     // Absolute y
     2 : int
     }
    // Number of armor points
    armor : int
    // Number of frames before next ground weapon possible attack
    gwcd : int
    // Ground weapon attack damage
    gwattack : int
    // Protoss shield points (like HP, but with special properties)
    shield : int
    // Air weapon attack damage
    awattack : int (air weapon attack damage)
    // Size of the unit
    size : int
    // Whether unit is an enemy or not
    enemy : bool
    // Whether unit is idle, i.e. not following any orders currently
    idle : bool
    // Ground weapon max range
    gwrange : int
    // Air weapon max range
    awrange : int
   }
  }
  // Same format as "units_myself"
  units_enemy : ...
 }
\end{lstlisting}



\renewcommand{\baselinestretch}{1}

\end{document}